\documentclass{article}
\usepackage{spconf,amsmath,graphicx}

\usepackage{subcaption}
\usepackage{multirow}
\usepackage{fancyhdr}

\usepackage{enumitem}
\setlist{nosep, leftmargin=14pt}

\usepackage{mwe} 

\def\air{1}
\def\mib{2}
\def\radis{3}
\def\uob{4}

\title{Location-based Radiology Report-Guided Semi-supervised Learning for Prostate Cancer Detection}
\name{
\begin{tabular}{@{}c@{}}
Alex Chen$^{\air}$ \qquad
Nathan Lay$^{\air}$ \qquad
Stephanie Harmon$^{\air}$ \qquad
Kutsev Ozyoruk$^{\air}$ \qquad \\
Enis Yilmaz$^{\mib}$ \qquad 
Brad J. Wood$^{\radis}$ \qquad
Peter A. Pinto$^{\uob}$ \qquad
Peter L. Choyke$^{\mib}$ \qquad \\
Baris Turkbey$^{\air\dagger}$\thanks{$^\dagger$ Corresponding author.}
\end{tabular}
}
\address{
$^{\air}$ Artificial Intelligence Resource, Molecular Imaging Branch, National Cancer Institute, Bethesda, MD \\
$^{\mib}$ Molecular Imaging Branch, National Cancer Institute, Bethesda, MD \\
$^{\radis}$ Radiology and Imaging Sciences, Clinical Center, National Institutes of Health, Bethesda, MD \\
$^{\uob}$ Urologic Oncology Branch, Center for Cancer Research, National Cancer Institute, Bethesda, MD
}
\fancypagestyle{FirstPage}{
\lfoot{\copyright2024 IEEE. Personal use of this material is permitted.  Permission from IEEE must be obtained for all other uses, in any current or future media, including reprinting/republishing this material for advertising or promotional purposes, creating new collective works, for resale or redistribution to servers or lists, or reuse of any copyrighted component of this work in other works.} 
}
\begin{document}
%
\maketitle
\begin{abstract}
\thispagestyle{FirstPage}
Prostate cancer is one of the most prevalent malignancies in the world. While deep learning has potential to further improve computer-aided prostate cancer detection on MRI, its efficacy hinges on the exhaustive curation of manually annotated images. We propose a novel methodology of semisupervised learning (SSL) guided by automatically extracted clinical information, specifically the lesion locations in radiology reports, allowing for use of unannotated images to reduce the annotation burden. By leveraging lesion locations, we refined pseudo labels, which were then used to train our location-based SSL model. We show that our SSL method can improve prostate lesion detection by utilizing unannotated images, with more substantial impacts being observed when larger proportions of unannotated images are used.

\end{abstract}
\begin{keywords}
Deep Learning, Semisupervised Learning, Prostate Cancer, MRI
\end{keywords}
\section{Introduction}
\label{sec:intro}
Prostate cancer remains one of the most prevalent malignancies worldwide, with estimates delineating it being the second most frequent cancer and the fifth leading cause of cancer death among men~\cite{sung2021global}. Use of biparametric magnetic resonance imaging (bpMRI) has become the standard for prostate cancer diagnosis, which has assisted with the early detection and treatment of localized disease~\cite{schroder2014screening}.
Nevertheless, diagnosis by bpMRI is highly dependent on radiologist expertise, leading to inter-observer differences and significant disparities between readers with different experience~\cite{girometti2019interreader}.

To address this challenge, computer-aided detection methods have received great focus, although automatic prostate cancer detection remains an incredibly challenging task. Multi-reader studies have shown that computer-aided detection methods can be misleading by detecting many false positives, leading to limitations in their translation to the clinic~\cite{greer2018computer}. 

Deep learning-based solutions have shown promise in lesion detection tasks for prostate MRI. However, developing a successful deep learning-based prostate cancer detection method requires a large amount of annotated data where the lesion is precisely segmented on the bpMRI. 
Supervised learning (SL) methods using tens of thousands of annotations can achieve the level of clinical experts~\cite{ardila2019end}, but acquiring the annotations is time intensive and costly, particularly for voxel-level segmentations. 
As a result, prostate cancer datasets have a limited number of precisely annotated images compared to the number of unannotated images. Other approaches are important to explore ways to reduce the annotation burden, utilize larger datasets, and ultimately develop a method with clinically translatable use for prostate cancer detection. 

To incorporate unlabeled medical imaging data, groups have utilized methods including self-supervised pretraining, transfer learning, and semisupervised learning (SSL) approaches~\cite{cheplygina2019not}. Since medical reports contain valuable clinical information and are available through clinical routine, Bosma et al utilized radiology reports by extracting the number of significant lesions in prostate MRIs to guide the generation of pseudo labels and build a state-of-the-art SSL model~\cite{bosma2023semisupervised}.
\begin{figure*}[t]
\captionsetup{font=small}
\centering
\includegraphics[width=17.79747cm]{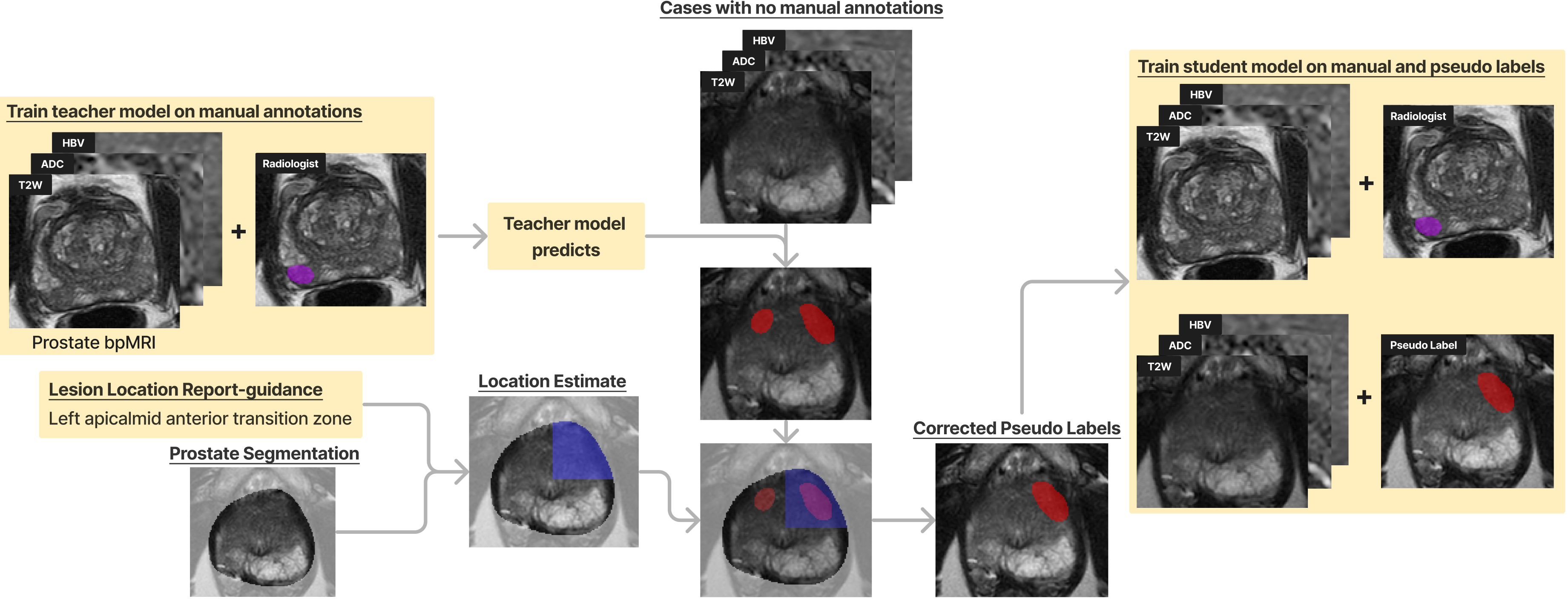}
\caption{A workflow of the semisupervised learning method with lesion location-based report-guidance for prostate cancer detection. bpMRI = biparametric MRI, T2W = T2-weighted, ADC = apparent diffusion coefficient, HBV = high-b value.}

\label{fig:workflow}
\end{figure*}
However, there is additional information in prostate radiology reports that can guide the generation of more robust pseudo labels for SSL. In addition, prostate segmentations are much more common than lesion segmentations due to their use in clinical routine for guided biopsies and therapy planning. To build upon~\cite{bosma2023semisupervised}, we devised an approach tailored for prostate lesion detection by automatically extracting the lesion locations found in a radiology report, which are then combined with a prostate segmentation to correct and generate pseudo labels. We compare our method with their report-guided SSL model and show that our lesion location-based SSL method can further improve the quality of pseudo labels, leading to overall improved detection by the model. 



\section{Materials \& Methods}
\label{sec:materials}
\subsection{Institutional Prostate MRI Dataset}


A cohort consisting of 2601 scans coming from 1837 patients (2015-2023) was queried from our institutional database. We used axial T2-weighted images, apparent diffusion coefficient (ADC) maps, and high b-value (HBV) images from each scan to train and evaluate models. The images were converted to NIFTI and resampled to have $0.5 \text{mm} \times 0.5 \text{mm} \times 3.0 \text{mm}$ voxel spacing. Images were normalized using z-score normalization. 
Among all scans, 1760 scans had annotations while 841 scans did not. Of the 841 unannotated scans, 619 contained clinically significant findings (PI-RADS $\geq 3$) in the radiology reports. For the SSL methods, we utilized 300, 500, and 1315 manually labeled cases, paired with the other 1634, 1434, and 619 unlabeled cases, respectively. The SL method utilized only the manually labeled cases. Models were trained with 5-fold cross-validation with randomly generated cross-validation splits and tested on the same set of 445 annotated cases.

\subsection{Lesion Location-based SSL}

Our SSL method follows a general workflow depicted in Fig.~\ref{fig:workflow}. Prostate lesion voxel-level segmentations were generated using the nnU-Net framework. The nnU-Net is a self-configuring framework that selects the preprocessing, network architecture, and post-processing steps~\cite{isensee2021nnu}. A teacher model was first trained on manually annotated images, which represents the SL method baseline. For the SSL methods, the teacher model was used to generate pseudo labels on unannotated cases by ensembling the predictions of the 5 models from 5-fold cross-validation, which are then corrected by using lesion locations from the radiology report. The pseudo labels are then combined with annotated cases to train a student model, which represents our lesion location-based SSL method. Our SSL method was compared with the SL model and the report-guided SSL method from~\cite{bosma2023semisupervised}, which was shown to perform better than other state-of-the-art SSL methods. Their SSL method will be referred to as the lesion count-based SSL method.

A clinically significant prostate lesion using MRI in the reports was defined as a lesion with PI-RADS $\geq 3$. Lesion locations were automatically extracted from radiology reports by utilizing the consistent structure between reports, allowing for the use of rule-based natural language processing. The location descriptions followed the convention detailed in the Sector Map of PI-RADS v2 and v2.1~\cite{turkbey2019prostate}. Utilizing the corresponding prostate segmentation, we used halved X and Y dimensions and a third of the Z dimension of the prostate to approximate location descriptions. Anatomical localizations (transition/peripheral zone) were ignored owing to not having transition or peripheral zone segmentations. After matching predicted lesions with the report-guided locations, matched lesions were kept and any unmatched lesions were removed, resulting in the pseudo labels for our SSL method. Cases with lesion locations that could not be matched to lesion predictions were removed. 


\subsection{Statistical Analysis}

\label{sec:statistics}
The performance of methods was measured with Dice Similarity Coefficient (DSC) and free-response Receiver Operator Characteristic (fROC). The DSC is a measure of segmentation performance and compares two segmentation masks.
The DSC produces a value in the range $[0,1]$ where $1$ implies the two masks are identical and $0$ implies that the masks have no voxels in common.

For lesion-based analysis, an fROC analysis was performed~\cite{bandos2019evaluation}. The 3D connected components were calculated on both the ground truth and segmentation mask, with the segmentation mask varying with probability threshold giving several points. Each ground truth connected component is considered a \textit{true positive} detection if the segmentation mask overlaps it, otherwise, the ground truth connected component is considered a \textit{false negative}. Each segmentation connected component is considered a \textit{false positive} if it does not overlap with a ground truth connected component. A strict detection for fROC analysis refers to lesions with greater than 0.1 Intersection over Union (IoU), which was only used to select a threshold where 60\% sensitivity was reached and to make pseudo labels from teacher model predictions. There is no measure of lesion-level \textit{true negatives} for our data set. As a consequence, there is no lesion-level ROC analysis.


\section{Results \& Discussion}
\label{sec:results}
\begin{table}[h]
\captionsetup{font=small}
\scriptsize
\centering
\begin{tabular}{c|c|c|c|c|c|c|}
\cline{2-7}
 & \multicolumn{6}{|c|}{Manually Labeled Training Cases} \\
\cline{2-7}
 & \multicolumn{2}{|c|}{300} & \multicolumn{2}{|c|}{500} & \multicolumn{2}{|c|}{1315} \\
\hline
\multicolumn{1}{|c|}{Pseudo label Correction} & TPR & FPpC & TPR & FPpC & TPR & FPpC \\
\hline
\multicolumn{1}{|c|}{No Correction} & 0.567 & 0.60 & 0.649 & 0.79 & 0.660 & 0.60 \\
\multicolumn{1}{|c|}{Count-based} & 0.675 & 0.41 & 0.704 & 0.41 & 0.736 & 0.36 \\
\multicolumn{1}{|c|}{Location-based} & 0.873 & 0.12 & 0.908 & 0.21 & 0.935 & 0.11 \\
\hline
\end{tabular}
\caption{Quality of pseudo labels evaluated by validation lesion-level false positives per case (FPpC) for Supervised models trained on 300, 500 and 1315 annotated cases with no correction, count-based corrections, or location-based corrections. Excluding cases with  lesion locations that could not be matched to predictions improved the sensitivity (TPR).}
\label{tbl:validperf}
\end{table}

\begin{figure}[t]
\captionsetup{font=small}
\centering
\includegraphics[height=2in]{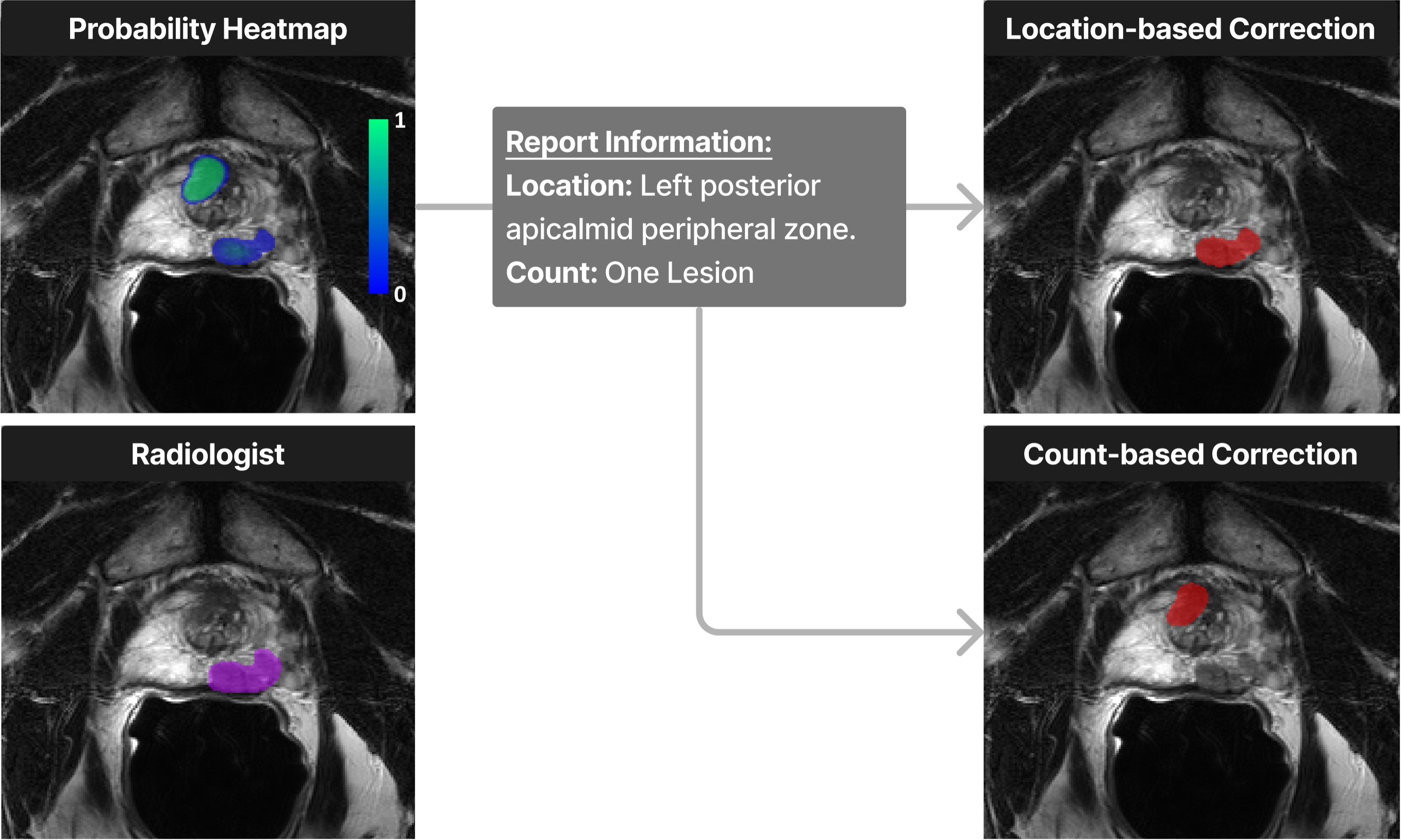}
\caption{Example case for when it is necessary to use lesion location-based report guidance to generate pseudo labels.}
\label{fig:pseudoMaskCountvsLoc}
\end{figure}




%

Ensuring high-quality pseudo labels is vital to this SSL method due to incorrect pseudo labels reinforcing incorrect teacher model predictions. Thus, the detection performance of the teacher model is examined on the validation set with or without report-guided corrections. In Table~\ref{tbl:validperf}, the teacher model achieves high sensitivity but many false positive lesions were also predicted. Although report-guided corrections based on lesion count reduced the false positive lesions per case, utilizing lesion location further reduced the false positives per case. For the teacher model trained on 1315 annotated cases, correcting lesion predictions based on lesion count reduced the mean false positive lesions per case from 0.60 to 0.36, which was further reduced to 0.11 by lesion location-based corrections. The reduction in false positive lesions greatly increased the quality of the pseudo labels. In addition, excluding cases with lesion locations that could not be matched to lesion predictions increased the sensitivity of the pseudo labels from 66.0\% to 93.5\%.

Fig.~\ref{fig:pseudoMaskCountvsLoc} reveals an example of when using location-based corrections is necessary. Given the probability heatmap of the teacher prediction, a count-based correction would keep the highest probability lesion and choose the incorrect one. In contrast, location-based corrections will keep the lesion that matches the location description, which guides the pseudo label to match the lesion identified by a radiologist. 

Bosma et al utilized lesion count-based corrections for PI-RADS $\geq$ 4 lesions while excluding PI-RADS 3 lesions~\cite{bosma2023semisupervised}. Since PI-RADS 3 lesions are less obvious but may still harbor clinically significant cancer, we included them in our dataset. This led to increased false positive lesion predictions by the teacher model when not corrected. As a result, by using PI-RADS $\geq$ 3 lesions, lesion location-based corrections were essential to reduce the false positives in pseudo labels and improve their quality.

\begin{figure}[t]
\captionsetup{font=small}
\centering
\includegraphics[height=2.4in]{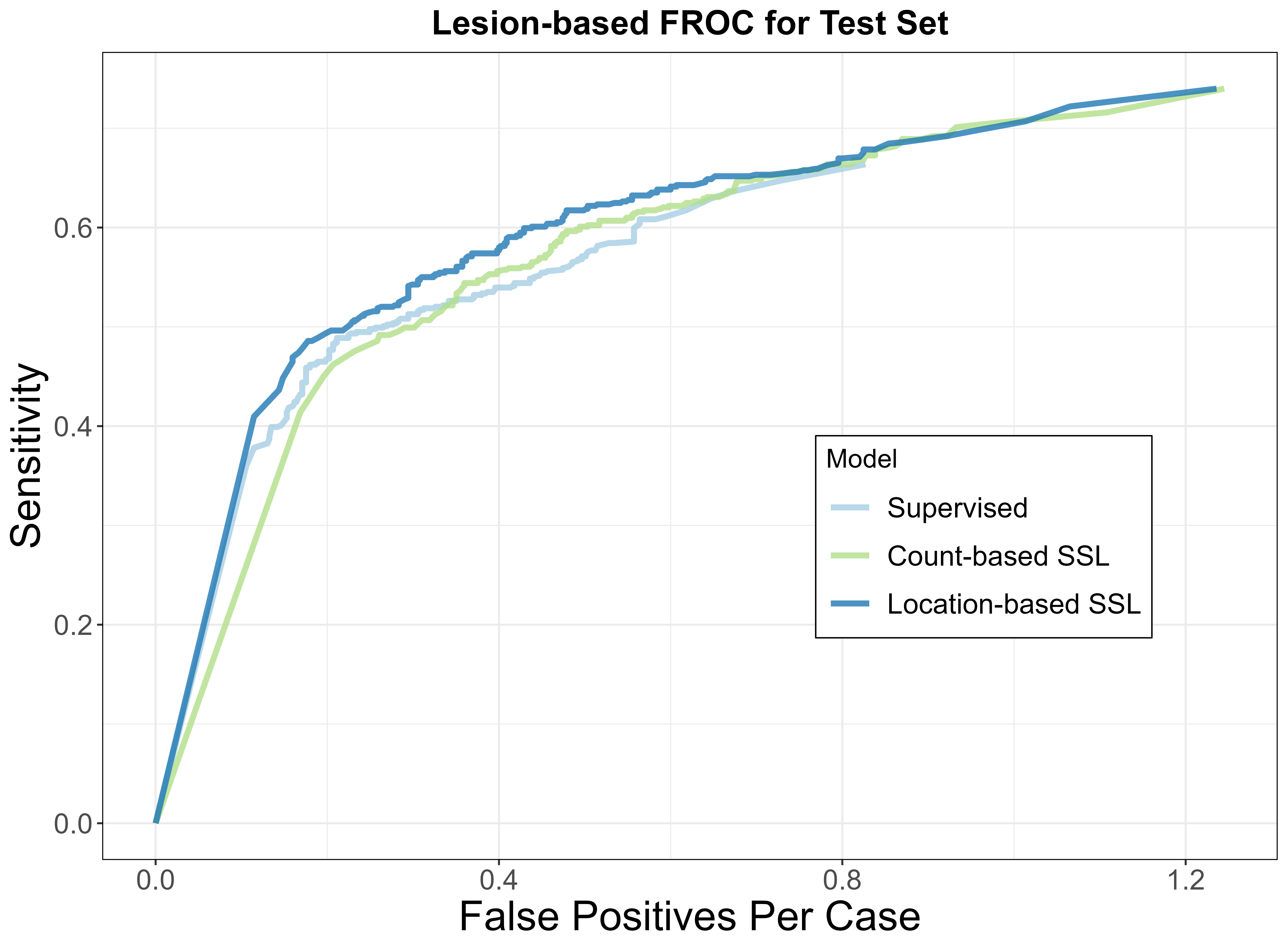}
\caption{Free-response receiver operating characteristic ROC (FROC) curve of the supervised model trained on 300 annotated cases compared to semi-supervised models trained on 300 annotated cases and 1634 unlabled cases and evaluated on 445 annotated test cases.}\label{fig:froc}
\end{figure}

The lesion location-based SSL model exhibits better detection performance than the supervised learning and lesion count-based SSL model. When 300 manually labeled cases were used, the fROC curve in Fig.~\ref{fig:froc} shows a consistent advantage for the lesion location-based SSL model. At this lower number of manually labeled cases, the ratio of pseudo labels to annotated cases is maximized and, therefore, the impact of the quality pseudo labels is most apparent. Moreover, the improved performance is due to the pseudo labels more accurately reflecting the radiology report.

\begin{table}[t]
\captionsetup{font=small}
\centering
\begin{tabular}{c|c|c|c|c}
\cline{2-4}
 & \multicolumn{3}{|c|}{Manually Labeled Cases} \\
\cline{1-4}
\multicolumn{1}{|c|}{Method} & 300 & 500 & 1315 \\
\hline
\multicolumn{1}{|c|}{Supervised Learning} & 0.62 & 0.58 & 0.35 & \multicolumn{1}{c|}{\parbox[t]{3mm}{\multirow{3}{*}{\rotatebox[origin=c]{-90}{FPpC}}}} \\
\multicolumn{1}{|c|}{Lesion Count SSL} & 0.49 & 0.47 & 0.33 & \multicolumn{1}{c|}{} \\
\multicolumn{1}{|c|}{Lesion Location SSL} & 0.44 & 0.44 & 0.30 & \multicolumn{1}{c|}{} \\
\hline
\end{tabular}
\caption{The false positives per case (FPpC) at 60\% sensitivity in the free-response receiver operating characteristic (FROC) curve evaluated on the test set for models trained on 300, 500, or 1315 manually labeled cases combined with 1634, 1434, or 619 unlabeled cases, respectively.}
\label{tbl:fppc}
\end{table}

Minimizing the number of predicted false positives per case while maintaining high sensitivity will be essential for using these automated systems to assist physicians in the future. Thus, the false positives per case at 60\% sensitivity were used to further analyze detection performance. At all number of manually labeled cases, the mean false positives per case at 60\% sensitivity is lowest for the lesion location-based SSL model, as shown in Table~\ref{tbl:fppc}. The detection performance showed greater improvements when fewer manually labeled cases were used to train the teacher model, likely due to the pseudo labels having a greater impact with a greater ratio of pseudo labels to manual annotations.

\begin{figure}[bt!]
\captionsetup{font=small}
\centering
\includegraphics[height=2.2in]{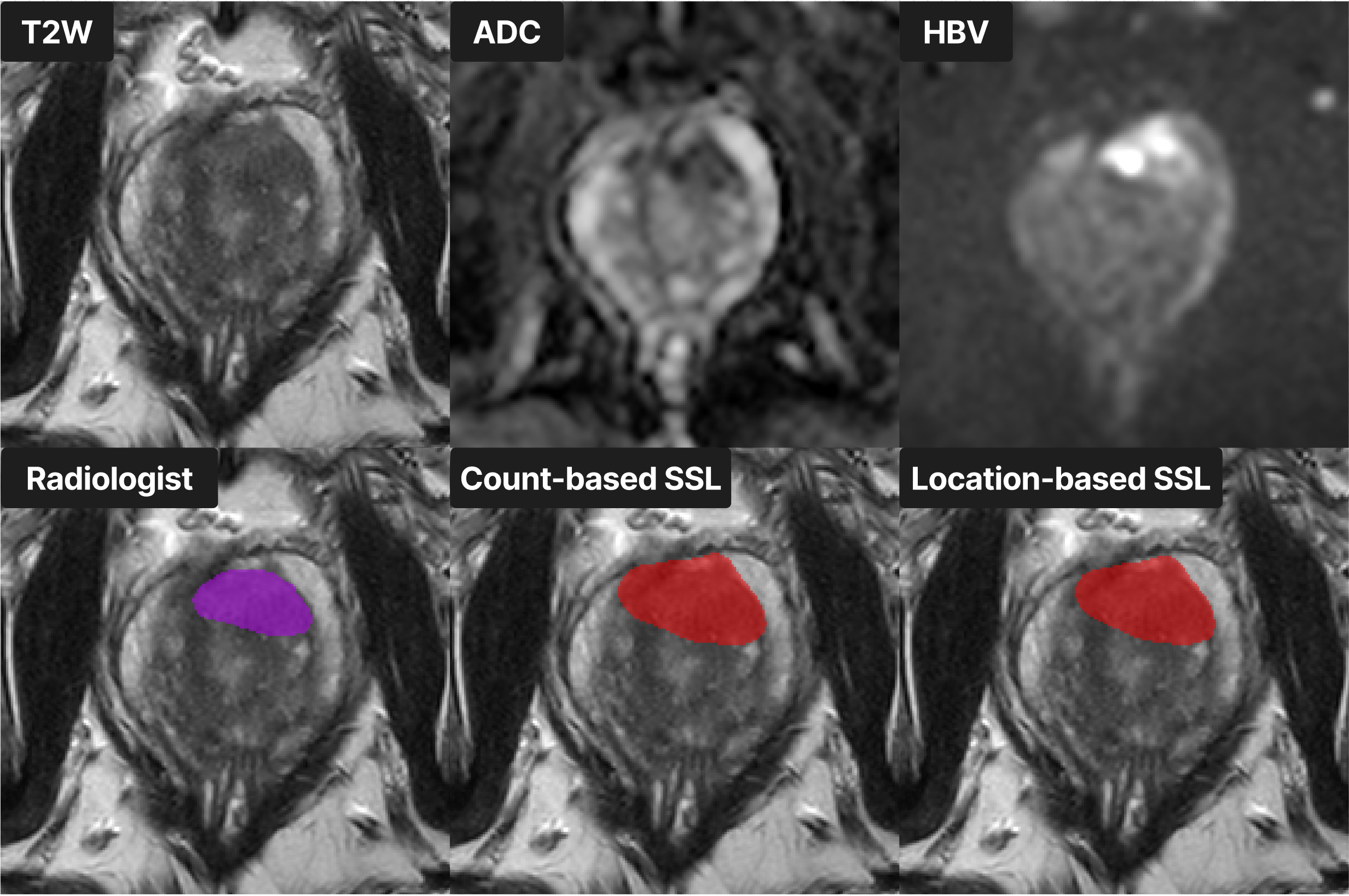}
\caption{Example patient from the test set with segmentations from the Count-based SSL (DSC=0.77) and Location-based SSL (DSC=0.82) trained on 500 manually labeled cases and 1434 unlabeled cases. T2W = T2-weighted, ADC = apparent diffusion coefficient, HBV = high-b value. }\label{fig:inferenceexample}
\end{figure}

The overall segmentation performance remains similar for supervised and semi-supervised learning for all experimental configurations as shown in Table~\ref{tbl:dsc}, although there are larger increases in DSC when fewer manually labeled cases are available. The similar DSCs are further hinted by Fig.~\ref{fig:inferenceexample} where the lesion count-based SSL and lesion location-based SSL models give visually near-identical segmentations.

Our lesion location-based SSL method has some limitations that should be considered. First, this SSL method relies on having large amounts of unannotated images. Our current dataset had limited amounts of unannotated images compared to annotated images. Thus, increasing the number of unannotated cases may lead to significant improvements in its prostate lesion detection, and allow us to realize the potential of our novel use of radiology reports.

\begin{table}[tb]
\captionsetup{font=small}
\centering
\begin{tabular}{c|c|c|c|c}
\cline{2-4}
 & \multicolumn{3}{|c|}{Manually Labeled Cases} \\
\cline{1-4}
\multicolumn{1}{|c|}{Method} & 300 & 500 & 1315 & \\
\hline
\multicolumn{1}{|c|}{Supervised Learning} & 0.334 & 0.366 & 0.424 & \multicolumn{1}{c|}{\parbox[t]{2mm}{\multirow{3}{*}{\rotatebox[origin=c]{-90}{DSC}}}} \\
\multicolumn{1}{|c|}{Lesion Count SSL} & 0.372 & 0.399 & 0.435 & \multicolumn{1}{c|}{} \\
\multicolumn{1}{|c|}{Lesion Location SSL} & 0.380 & 0.395 & 0.434 & \multicolumn{1}{c|}{} \\
\hline
\end{tabular}
\caption{DSC scores on the test set for the supervised model trained on 300, 500, or 1315 manually labeled cases compared to semi-supervised models utilizing an additional 1634, 1434, 619 unlabeled cases, respectively.}
\label{tbl:dsc}
\end{table}


The rule-based extraction of lesion locations relies on the structure of the report, restricting its use in unstructured reports. Recent findings have shown the impressive ability of large language models to convert unstructured reports to structured reports, although careful evaluation of use on new reports may be needed~\cite{adams2023leveraging}. Moreover, with the non-uniform nature of annotations, there is similar heterogeneity in reports, leading to a uniformity challenge that may limit this method when using lower-quality reports. 

Lastly, since the use of lesion location-based report guidance relied on standardized prostate locations described in PI-RADS, this method is limited in applicability to mainly other malignancy detection models that have consistent location descriptions, such as lung, liver, and brain malignancies.



\section{Conclusion}
\label{sec:conclusion}
In this paper, we proposed an innovative method of utilizing valuable information from radiology reports for SSL in prostate cancer detection. Utilizing lesion location automatically extracted from radiology reports generates pseudo labels with few errors. While test DSCs remained similar compared to other approaches, the false positive rate was reduced when using lesion location-based pseudo labels, particularly when greater proportions of pseudo labels were incorporated. Our approach allows for a reduction of the annotation burden with a novel approach to using radiology reports, opening a path forward for the use of substantially larger datasets in prostate cancer detection models. 


\section{Compliance with Ethical Standards}
\label{sec:ethics}
This single-institution study was approved by the NIH institutional review board. Written informed consent was obtained from all patients.

\section{Acknowledgments}
\label{sec:acknowledgments}
This research was funded by the NIH intramural research program. This work utilized the computational resources of the NIH HPC Biowulf cluster. (http://hpc.nih.gov)

\bibliographystyle{IEEEbib}
\bibliography{paper}

\end{document}